\documentclass{article}
\usepackage{PRIMEarxiv}

\usepackage[T1]{fontenc}
\usepackage[utf8]{inputenc}
\usepackage{lmodern}
\usepackage{microtype}

\usepackage{amsmath, amssymb, amsfonts, mathtools}

\usepackage{graphicx}
\graphicspath{{media/}}
\usepackage{booktabs}
\usepackage{caption}
\usepackage{float}

\usepackage{enumitem}
\setlist[itemize]{noitemsep, topsep=2pt}
\setlist[enumerate]{noitemsep, topsep=2pt}

\usepackage[authoryear,round]{natbib}

\usepackage{xurl}
\usepackage[hidelinks]{hyperref}
\hypersetup{
  pdftitle={One Global Model, Many Behaviors},
  pdfauthor={Bartosz Szab\l{}owski},
  bookmarksnumbered=true
}
\usepackage[noabbrev,nameinlink]{cleveref}

\usepackage{tikz}
\usetikzlibrary{positioning,fit,arrows.meta,calc}

\usepackage{fancyhdr}
\title{One Global Model, Many Behaviors\\
\large Stockout-Aware Feature Engineering and Dynamic Scaling for Multi-Horizon Retail Demand Forecasting\\
\large with a Cost-Aware Ordering Policy (VN2 Winner Report)}

\author{Bartosz Szabłowski}

\begin{document}
\maketitle

\begin{abstract}
Inventory planning for retail chains requires translating demand forecasts into ordering decisions, including asymmetric shortages and holding costs. The VN2 Inventory Planning Challenge formalizes this setting as a weekly decision-making cycle with a two-week product delivery lead time, where the total cost is defined as the shortage cost plus the holding cost. This report presents the winning VN2 solution: a two-stage predict-then-optimize pipeline that combines a single global multi-horizon forecasting model with a cost-aware ordering policy. The forecasting model is trained in a global paradigm, jointly using all available time series. A gradient-boosted decision tree (GBDT) model implemented in CatBoost is used as the base learner. The model incorporates stockout-aware feature engineering to address censored demand during out-of-stock periods, per-series scaling to focus learning on time-series patterns rather than absolute levels, and time-based observation weights to reflect shifts in demand patterns. In the decision stage, inventory is projected to the start of the delivery week, and a target stock level is calculated that explicitly trades off shortage and holding costs. Evaluated by the official competition simulation in six rounds, the solution achieved first place by combining a strong global forecasting model with a lightweight cost-aware policy. Although developed for the VN2 setting, the proposed approach can be extended to real-world applications and additional operational constraints.
\end{abstract}

\keywords{time series \and inventory planning \and demand forecasting \and global forecasting model \and CatBoost \and gradient boosting decision trees \and multi-horizon forecasting \and feature engineering \and retail time series \and cost-aware ordering policy}

\section{Introduction}
\label{sec:introduction}
Retail inventory planning combines demand forecasting with operational decision-making. A forecast has limited value on its own unless it is translated into ordering decisions that balance product availability against excess inventory. In retail, this trade-off is especially delicate due to the large number of SKUs, highly heterogeneous demand patterns across products and stores, and the business impact of stockouts. The VN2 Inventory Planning Competition provides a clear and reproducible benchmark for this problem by requiring weekly ordering decisions under a fixed two-week lead time, while minimizing a total cost composed of shortage costs (lost sales) and holding costs \citep{vn2challenge2024}.

The VN2 setting exposes three practical challenges in translating forecasts into inventory decisions, which often separate “good forecasts” from “good ordering decisions”.
First, time series vary across products; even the same product can exhibit different patterns across stores. In practice, data may include regular versus intermittent demand, trends or no trend, seasonality or no seasonality, and behaviors that change over time, with series differing in scale. This matters because aggregate error metrics such as WAPE or BIAS can implicitly prioritize large-volume series while underweighting small or intermittent ones, which complicates both model selection and learning objectives. Moreover, local (per-series) approaches can be costly to maintain and tune at scale, while global approaches typically require advanced design but provide value through cross-learning \citep{salinas2020deepar,smyl2020hybrid,joseph2022mtsfp}.
Second, time series are often censored by on-shelf availability: observed sales depend on whether inventory was available, so periods of constrained availability do not reveal true underlying demand \citep{lariviere1999stalking}. Training a forecasting model directly on observed sales can therefore introduce downward bias and lead to systematic underestimation, especially for fast movers or promotional-like spikes that were truncated by stockouts.
Third, decision-making must explicitly account for lead time and asymmetric costs. With a two-week delay between ordering and delivery, an order decision is not about replenishing today’s shelves, but about positioning inventory for future sales weeks, while trading off shortage and holding costs.

This report describes the winning VN2 solution, structured as a two-stage predict-then-optimize pipeline. The first stage is a global multi-horizon forecasting model based on CatBoost. To achieve this, the time-series problem is transformed into a tabular regression task and used to train a global model \citep{januschowski2020criteria}. The forecasting stage incorporates stockout-aware feature engineering, which is grouped into: (i) behavioral features capturing heterogeneous demand patterns, (ii) per-series dynamic scaling to emphasize patterns rather than absolute volume levels, and (iii) time-decayed observation weighting to adapt to changes in behavior over time. The second stage converts forecasts into ordering decisions by projecting inventory to the delivery week and computing a cost-aware target stock level, resulting in a simple, robust, and interpretable rule aligned with the competition objective.

The goal of this report is not only to document the winning solution, but also to extract practical design principles for supply chain and demand forecasting. While the VN2 competition imposes a specific setup, the proposed pipeline is intentionally modular and can be extended to other settings and operational constraints.
The remainder of this report is organized as follows: Section 2 describes the problem setup and simulation rules; Section 3 presents the data; Section 4 presents existing approaches; Sections 5–7 describe the two-stage solution: demand forecasting and the ordering policy; and Section 8 concludes with key lessons and directions for extensions.

\section{Problem Setup}
\label{sec:problem}
VN2 is a periodic-review inventory planning problem for grocery retail. Time is discretized into weekly periods. For each item (store--product pair) $i \in \{1,\dots,N\}$ with $N=599$, the order is an integer $Q_{t}^{(i)} \ge 0$ once per week. Orders are placed at the end of week $t$ (before week $t+1$ starts) and delivered after a fixed lead time of two weeks, so an order placed at the end of week $t$ becomes available for sales at the start of week $t+3$.

\subsection{Inventory dynamics and lost sales}
Let $I_{t}^{(i)}$ denote on-hand inventory at the start of week $t$ (after receiving scheduled deliveries). For week $t$, demand $D_{t}^{(i)}$ is realized as sales $S_{t}^{(i)}$ up to the on-hand inventory $I_{t}^{(i)}$ available in that period:
\begin{align}
S_{t}^{(i)} &= \min\{I_{t}^{(i)},\, D_{t}^{(i)}\}.
\end{align}
Lost sales $L_{t}^{(i)}$, also called shortages, occur whenever demand $D_{t}^{(i)}$ is higher than the inventory available $I_{t}^{(i)}$:
\begin{align}
L_{t}^{(i)} &= \max\{D_{t}^{(i)} - I_{t}^{(i)},\, 0\}.
\end{align}
There are no backorders, so it does not carry over to future periods.\\
End-of-week inventory is
\begin{align}
E_{t}^{(i)} = I_{t}^{(i)} - S_{t}^{(i)}.
\end{align}
Excess inventory is carried over to the next period; there is no risk of waste, loss, or obsolescence.

To represent the delivery pipeline, we maintain in-transit orders scheduled to arrive in future weeks. Let $R_{t}^{(i)}$ be the quantity received at the start of week $t$ (orders placed at the end of week $t{-}3$). The start-of-week inventory update is:
\begin{align}
I_{t+1}^{(i)} = E_{t}^{(i)} + R_{t+1}^{(i)}.
\end{align}

\subsection{Objective function}
The goal is to minimize the total cost over the simulation horizon:
\begin{align}
\min_{Q_{1:T}} \ \sum_{t=1}^{T}\sum_{i=1}^{N}
\left(
c_s \, L_{t}^{(i)} + c_h \, E_{t}^{(i)}
\right),
\end{align}
with shortage cost $c_s = 1.0$ per unit of lost sales and holding cost $c_h = 0.2$ per unit of end-of-week on-hand inventory. No holding cost is applied to in-transit goods. Supply is assumed to be unconstrained with stable lead times.

\subsection{Competition setup}
At the beginning of each decision period, the available information consists of the current inventory state (on-hand and in-transit deliveries), the full history of weekly sales, and records of historical stockout periods. Participants submit one order quantity per store–product series each week for six consecutive weeks (Rounds 1--6). Due to the two-week lead time, the simulator continues for two additional weeks (up to Week 8) to account for the final delivery. All participants start from an identical initial state and face the same demand during the competition.

\section{Data}
\label{sec:data}
The VN2 Challenge data is organized at a weekly granularity for a fixed set of $N=599$ store--product pairs (67 unique stores and 297 unique products). At the beginning (Week~0), participants receive: (i) historical weekly sales, (ii) a binary on-shelf availability information (\textit{in-stock}), (iii) static master data describing stores and products, and (iv) an initial inventory snapshot including on-hand and in-transit deliveries. During the competition rounds, the only new information is the most recent week of demand that occurred, appended to the sales history.

Historical sales are provided in a wide format with one column per week indicated as the 1st day (Monday) of each week. The initial history contains 157 weeks (~3 years) from 2021-04-12 to 2024-04-08; after each round, the next week becomes available in the updated sales file.

\begin{figure}[H]
    \centering
    \includegraphics[width=\linewidth]{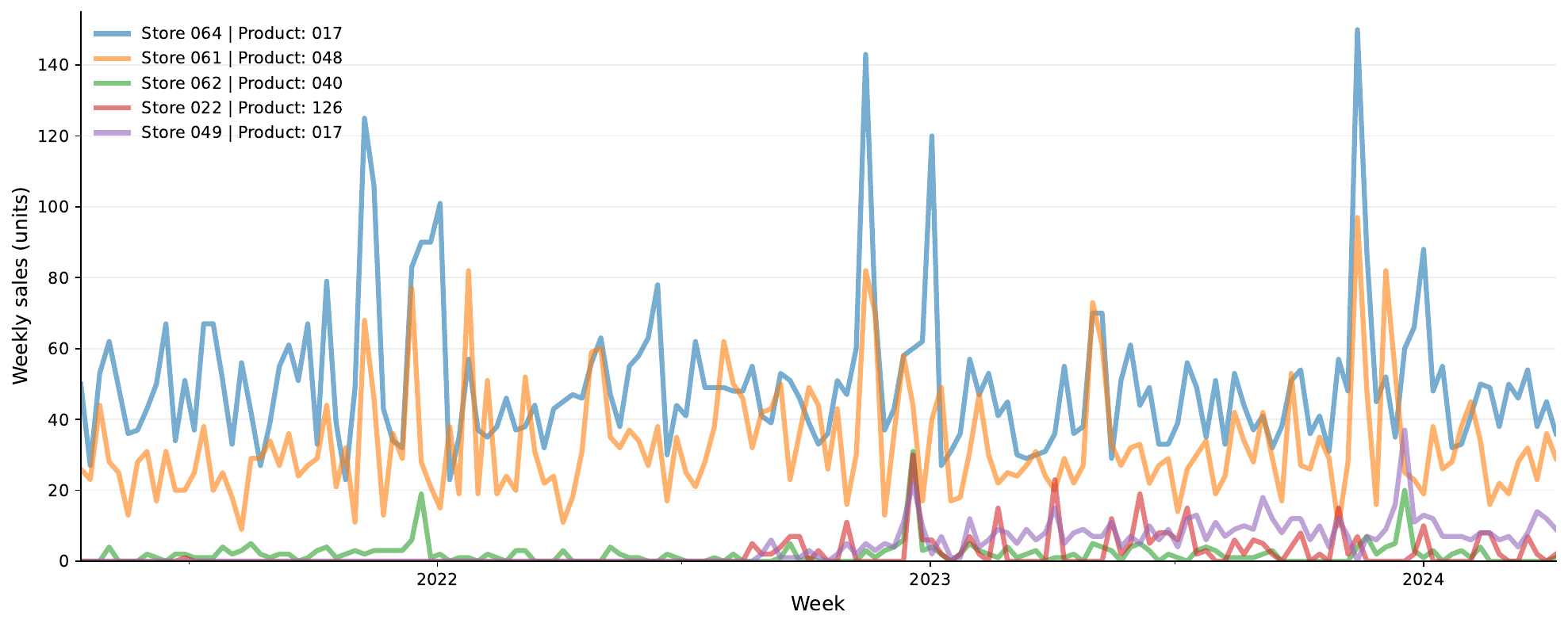}
    \caption{Examples of five store--product weekly sales series illustrating differences in scale, intermittency, seasonality, and delayed starts.}
    \label{fig:vn2_ts_examples}
\end{figure}

The dataset exhibits substantial diversity in time-series characteristics. Across all store--product histories, approximately 43\% of weekly observations have zero sales, indicating frequent intermittency \citep{croston1972intermittent,syntetos2009review}. Even if the entire history is about 3 years, not all time series begin with sales at the beginning of history. Series also differ strongly in scale: average weekly sales range from roughly 0.3 to nearly 99 units per week across items. In-stock availability is high on average, but stockouts are non-negligible (about 10.6\% of weekly observations are marked as not in stock), and whenever \texttt{In Stock} is \texttt{False}, observed sales are zero. As a result, the sales history should be interpreted as constrained rather than unconstrained demand. Figure~\ref{fig:vn2_ts_examples} provides five example series highlighting these differences in time-series behavior.

\section{Existing Inventory Planning Approaches for Order Optimization}
\label{sec:methodology_approaches}

In inventory planning, a two-stage predict-then-optimize workflow is the common operational paradigm: demand is first forecasted, and the forecast is then translated to an ordering decision. Classical periodic-review policies (e.g., order-up-to / base-stock logic) combine an estimate of demand over the protection period (lead time plus review period) with a safety buffer chosen to balance holding and shortage costs \citep{silver1998inventory,zipkin2000foundations}. In the single-period approximation, this trade-off reduces to the newsvendor critical fractile, as it targets the demand quantile implied by asymmetric costs.

A key limitation of an approach that focuses solely on forecast accuracy is that improvements in standard error metrics do not necessarily translate into lower inventory costs, especially under intermittent demand, stockout-censored observations, and lead-time aggregation. Even when the underlying demand process is weakly autocorrelated, multi-step forecast errors over the lead time can be positively correlated, which directly affects safety stock and reorder level computations \citep{prak2017safetystock}. As a result, many applied systems retain the two-stage structure, but place emphasis on (i) robust forecasting under retail data issues and (ii) a cost-consistent mapping from forecasts (and their uncertainty) to stock targets.

Beyond point forecasts, a widely used extension is probabilistic forecasting and scenario-based planning: safety buffers are derived from predictive distributions (or simulated paths) of lead-time demand rather than from Gaussian approximations. When uncertainty is multi-period or multi-item, dependence across forecast errors becomes important. Copula models provide a flexible way to construct a joint error (or demand) distribution by coupling marginal predictive distributions with an explicit dependence structure \citep{nelsen2006copula,patton2012copula,silbermayr2017copula}. In practice, such joint modeling can be used to sample correlated error scenarios over the protection period and evaluate candidate ordering policies under realistic uncertainty.

Alongside these extensions of the two-stage paradigm, research has developed more integrated approaches that couple prediction with downstream decisions. In the conventional predict-then-optimize paradigm, a predictive model is trained first (often for a statistical error), and a separate downstream policy then maps predictions into decisions and costs. In predict-and-optimize, the predictive
model is trained by directly minimizing the downstream decision cost, so the decision problem enters the learning objective \citep{elmachtoub2022spo,donti2017e2e,bertsimas2020prescriptive,vanderschueren2022predict}. Related streams study data-driven newsvendor variants that learn order quantities directly from contextual features, bypassing explicit distribution fitting while still targeting asymmetric costs \citep{ban2018bigdata,huberv2019newsvendor}. For multi-period settings with lead times, capacity constraints, and lost sales, reinforcement learning has also been explored, but typically requires careful simulation design and stable policy optimization \citep{boute2022drlroadmap,madeka2022deepinventory,alvo2023hdpo}.

\section{Overall Solution: Two-Stage Approach}
\label{sec:overall}
The proposed solution follows a two-stage \textit{predict-then-optimize} approach. The key idea is to decouple (i) demand forecasting from (ii) generating orders decisions under lead time and asymmetric costs. This separation results in a scalable and modular pipeline. Components can be developed and improved independently, while acknowledging that the order decision depends solely on forecast quality. In practice, this modularity simplifies iteration and validation. It becomes clear whether performance gains require better demand estimates or a better ordering policy, rather than treating the system as an opaque end-to-end block.

\begin{figure}[tb]
\centering
\begin{tikzpicture}[
  font=\small,
  box/.style={
    draw, rounded corners=2pt, align=left,
    inner sep=6pt, text width=4.3cm, minimum height=1.35cm
  },
  mdl/.style={
    draw, rounded corners=2pt, align=left,
    inner sep=7pt, text width=4.7cm, minimum height=1.55cm
  },
  arr/.style={-Latex, thick},
  stage/.style={draw, rounded corners=2pt, dashed, thin, inner sep=10pt},
  title/.style={font=\bfseries\small}
]

\node[box] (hist) {\textbf{Historical Demand\\\& Availability}\\
\(\bullet\) Weekly sales history\\
\(\bullet\) In-stock availability flag};

\node[mdl, right=14mm of hist] (forecast) {\textbf{Demand Forecasting Model}\\
\(\bullet\) Global multi-horizon forecaster\\
\(\bullet\) Point forecasts (\(t{+}1, t{+}2, t{+}3\))};

\node[box, right=14mm of forecast] (fc_out) {\textbf{Forecasted Demand}\\
\(\hat{D}_{t+1}, \hat{D}_{t+2}, \hat{D}_{t+3}\)};

\draw[arr] (hist) -- (forecast);
\draw[arr] (forecast) -- (fc_out);

\node[box, below=16mm of hist] (state) {\textbf{Current State}\\
\(\bullet\) On-hand inventory\\
\(\bullet\) In-transit \(t{+}1, t{+}2\)};

\node[mdl, right=14mm of state] (policy) {\textbf{Ordering Policy}\\
\(\bullet\) Inventory projection to \(t{+}3\)\\
\(\bullet\) Cost-aware target stock\\
\(\bullet\) Order rule};

\node[box, right=14mm of policy] (orders) {\textbf{Orders for Week \(t\)}\\
targeting availability at \(t{+}3\)};

\draw[arr] (state) -- (policy);
\draw[arr] (policy) -- (orders);

\draw[arr] (fc_out.south) -- ++(0,-6mm) -| (policy.north);

\node[stage, fit=(hist)(forecast)(fc_out)] (stage1) {};
\node[stage, fit=(state)(policy)(orders)] (stage2) {};

\node[title, above=1mm of stage1] {Stage 1};
\node[title, below=4mm of policy] {Stage 2};

\end{tikzpicture}
\caption{Two-stage predict-then-optimize pipeline. A global multi-horizon forecasting model produces demand forecasts for \(t{+}1\) to \(t{+}3\). The ordering policy combines these forecasts with the current inventory state to compute a cost-aware order decision targeting availability in week \(t{+}3\).}
\label{fig:two_stage_pipeline}
\end{figure}
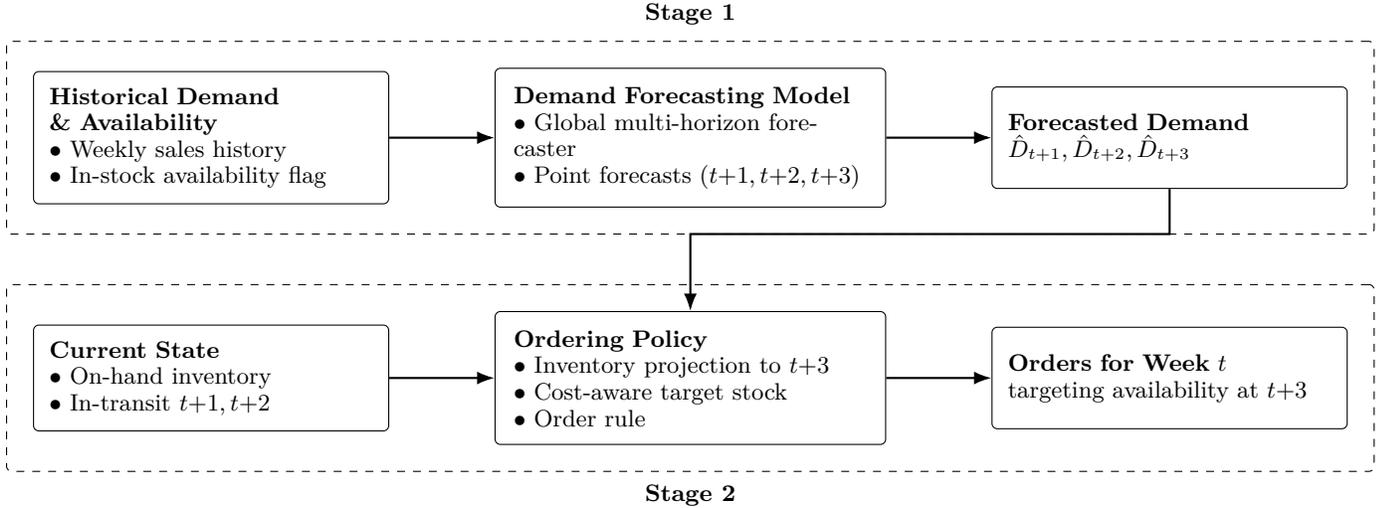

\paragraph{Stage 1: Global multi-horizon demand forecasting.}
In the first stage, a single global model is trained that produces point forecasts for the next three weeks ($t{+}1$ to $t{+}3$) for each store--product pair. The model incorporates historical weekly sales together with availability information (in-stock flag). A global approach is particularly suitable in this setting because the scope contains many heterogeneous series with varying scale and intermittency, making local per-series tuning costly, brittle, and hard to maintain. Evidence from large-scale forecasting competitions (like M5) suggests that carefully designed global, feature-based machine learning approaches, often leveraging cross-series learning can achieve state-of-the-art accuracy on heterogeneous retail demand data \citep{makridakis2022m5}. The three-week horizon is chosen to match the operational lead time: orders placed at week $t$ only impact availability in week $t{+}3$, while intermediate forecasts ($t{+}1$, $t{+}2$) are used to project inventory at the start of $t{+}3$.

\paragraph{Stage 2: Cost-aware ordering policy.}
The second stage converts forecasts into an order quantity for week $t$, targeting inventory availability in week $t{+}3$. The policy combines forecasted demand with the current inventory state (on-hand and in-transit) to project the inventory position at the start of the delivery week and to compute a target stock level that explicitly trades off shortage and holding costs. This stage is lightweight, but directly optimizes the same cost structure used for evaluation, while leveraging the forecasting model as a strong demand signal. Also, it is possible to extend to different perspectives to better tailor it to new cases.

\paragraph{Why two stages?}
This design choice is motivated by three considerations. First, it improves \textit{tractability and stability}: demand forecasting can be trained with standard supervised learning objectives, while inventory decisions can be derived via an explicit cost-aware mapping. Second, it improves \textit{scalability}: a single global forecaster and a simple policy can be applied consistently across hundreds of items without per-series calibration. Third, it improves \textit{interpretability and extensibility}: the forecasting model and the ordering policy can be analyzed and validated separately, and extended independently (e.g., incorporating additional operational constraints, including new patterns in time series forecasting, etc.).

Figure~\ref{fig:two_stage_pipeline} summarizes the overall information flow. Detailed descriptions of the forecasting model and the ordering policy are provided in Sections~\ref{sec:forecasting_model} and~\ref{sec:ordering_policy}, respectively.

\section{Forecasting Model}
\label{sec:forecasting_model}
The forecasting component is the backbone of the proposed predict-then-optimize pipeline. The ordering policy is lightweight (Section~\ref{sec:ordering_policy}) and its performance is driven by forecast accuracy. This section describes (i) the global paradigm, multi-horizon strategy, training setup, and (ii) the feature engineering design that enables a single tabular estimator to generalize across many store--product patterns.

\subsection{Global Forecasting Model}
\label{sec:global_forecasting_model}

\paragraph{Global forecasting models.}
A forecasting model is \textit{global} when a single estimator is trained on multiple time series and can generate forecasts for all series, rather than being fitted to a single, specific series (a local approach). This is particularly well-suited to the VN2 setting because the portfolio contains many heterogeneous series (in scale and behavior), and local per-series approaches would require extensive tuning. In contrast, a global model can borrow information across series and learn overarching demand patterns \citep{salinas2020deepar,smyl2020hybrid,hyndman2021fpp3}.

\paragraph{Supervised learning formulation.}
Because true demand $D_t^{(i)}$ is censored during stockouts, the model is trained to predict an \emph{effective demand proxy} constructed from observed sales in in-stock weeks and treated as missing otherwise; I denote its forecast by $\hat{D}_{t+h}^{(i)}$ for simplicity. The time-series forecasting task is cast as a supervised learning problem by converting the sequential data into a tabular dataset. For each item–week observation, a feature vector is constructed using only information available up to that week, including created features that encode past values and categorical or calendar information into a compact numerical representation suitable for supervised learning. The target for each row is a vector of future demands(t+1,t+2,t+3) \citep{bontempi2013mlstrategies, joseph2022mtsfp}.

\paragraph{Multi-horizon forecasting via the direct strategy.}
The Forecasting horizon is a three-week ($t{+}1$, $t{+}2$, $t{+}3$), and it is adopted as \textit{direct} multi-horizon strategy: a separate estimator is trained for each horizon $h \in \{1,2,3\}$ to predict $\hat{D}_{t+h}^{(i)}$ based on features at time $t$. This is implemented by shifting the target forward by $h$ steps within each series and fitting three independent estimators. Alternative strategies include \textit{recursive} forecasting (predict one step ahead and feed predictions back as inputs for longer horizons) and \textit{joint strategy} forecasting (a single model predicting a vector of horizons). In addition, hybrid approaches such as DirRec, IBD, Rectify, and RecJoint are also known \citep{joseph2022mtsfp}. Recursive approaches can accumulate error across steps and are sensitive to distribution shift between training and inference. The joint strategy trains a single estimator to predict the full-horizon vector in a single forward pass, $[\hat{y}_{t+1},\ldots,\hat{y}_{t+H}]$, rather than producing one-step-ahead forecasts and feeding them back recursively.
While this can avoid error accumulation and enforce cross-horizon consistency, it introduces additional design choices: how parameters are shared across horizons, how the joint loss is defined. In practice, such multi-output setups are most common in deep-learning models. As a result, it has a higher entry barrier than direct or recursive strategies, which are easier to implement. In practice, many teams prefer direct or recursive approaches because, despite their simplicity, they are often sufficient to achieve strong performance. In VN2, the horizon is short (only three steps), and the ordering decision depends primarily on $t{+}3$; thus, the direct strategy is a pragmatic choice: it avoids error accumulation, allows horizon-specific modeling, and remains computationally tractable \citep{bontempi2013mlstrategies}.

\paragraph{Estimator choice: CatBoost.}
CatBoost is used as the base learner. The choice is motivated by the following considerations. First, the forecasting task is expressed as tabular regression with a mix of numeric features (lags, rolling statistics, seasonal terms) and categorical features (store/product/series ID and week of year). It provides native support for categorical variables, reducing the need to manually encode static covariates. In the competition setting, series identifiers were intentionally used as categorical features to improve the leaderboard metric. In production forecasting systems, such identifiers typically do not generalize to new unseen stores or products and may require frequent retraining, so more stable parent-level categories or other transferable representations are usually preferred. Second, as a tree-ensemble method, it naturally captures non-linear interactions between engineered features (like intermittency indicators interacting with seasonal proxies, trend combined with seasonality signals, or seasonality features modulated by item-specific sensitivity) \citep{januschowski2022forecastingtrees}. It is an important property when a single global model must represent heterogeneous demand behaviors across many time series. Finally, CatBoost was validated and compared with other strong estimators, such as Random Forest \citep{breiman2001randomforests} and LightGBM \citep{ke2017lightgbm}. The comparison used the same engineered features, the same validation set, and a comparable optimization budget across estimators. However, that performance is tightly coupled to the feature engineering design; the same estimator can yield materially different results under alternative feature representations \citep{prokhorenkova2018catboost}.

\paragraph{Hyperparameter optimization and time-based validation.}
Data are first split into a training set and a local time-based test holdout (the last 18 weeks). Within the training set, a chronological split creates a training subset and a validation subset used for hyperparameter selection. Concretely, the most recent 10\% of weeks within the training set is reserved for validation, ensuring that candidates are evaluated on future observations relative to fitting. For each horizon-specific estimator, Optuna runs 100 trials with a TPE (Tree-structured Parzen Estimator) sampler, a Bayesian optimization strategy, and searches over key CatBoost hyperparameters (depth, learning rate, $L_2$ regularization, feature subsampling, bootstrap type, and its associated hyperparameters, etc.). Estimators are trained with a large iteration budget and early stopping (500 rounds) to prevent overfitting, and hyperparameters are selected based on validation performance on this internal split \citep{akiba2019optuna}. Using the chosen hyperparameters, the model is then refit on the full training set and evaluated via a backtest on the local test holdout (18 weeks) for performance assessment only. Finally, keeping the hyperparameters fixed, the model is trained on the full available history and used to generate the predictions for the final submission. Each CatBoost regressor is trained using the RMSE loss on the scaled target, while Optuna selects hyperparameters by minimizing a validation MAE computed after inverting predictions back to the original sales units. This choice is intentional and reflects a practical trade-off between stable global fitting and business relevance. RMSE is used as the CatBoost training loss on the scaled target to prevent the level of series from dominating optimization, while Optuna evaluates candidates using a MAE in the original sales units, so that higher-throughput products carry proportionally more influence in hyperparameter selection. MAE is less sensitive to occasional demand spikes, which makes the hyperparameter selection more stable.

\subsection{Feature Engineering}
\label{sec:feature_engineering}

\paragraph{Behavioral features.}
Features were added iteratively. Starting with simple features such as lag features and moving averages, and then analyzing where the estimator makes the biggest mistakes: which situations and patterns it forecasts the worst, what drives those errors, and what information is missing. Based on that analysis, the next set of features was introduced and evaluated to confirm that it added value. These created features we can split into groups that describe (i) demand level and recent history, (ii) trend and momentum, (iii) seasonality, and (iv) intermittency and spike behavior, while explicitly accounting for stockout censoring. To mitigate distorted data during out-of-stock weeks, an effective sales series $y^{\mathrm{eff}}$ is defined by masking weeks with $\texttt{in\_stock}=\texttt{False}$, and target-based features are computed on $y^{\mathrm{eff}}$ rather than on raw sales:
\[
y_{t}^{\mathrm{eff},(i)} =
\begin{cases}
y_{t}^{(i)}, & \text{if } \texttt{in\_stock}_{t}^{(i)}=\texttt{True},\\[2pt]
\mathrm{NaN}, & \text{if } \texttt{in\_stock}_{t}^{(i)}=\texttt{False}.
\end{cases}
\]
The resulting feature set includes short-term lags ($t, t{-}1, t{-}2, t{-}3$) and seasonal lags ($t{-}51, t{-}52, t{-}53$), rolling means and medians (windows 3, 5, 13), exponentially weighted means (spans 5 and 10), and dispersion measures (rolling standard deviation and interquartile range). 

Trend and momentum are represented by short-horizon differences and a rolling-slope proxy.
Momentum is computed as $\Delta_k y_t^{\mathrm{eff}} = y_t^{\mathrm{eff}} - y_{t-k}^{\mathrm{eff}}$ for $k\in\{1,5\}$, while the slope proxy is defined as the moving average of first differences over a 4-week window,
$\mathrm{slope}_t = \frac{1}{4}\sum_{j=0}^{3}\left(y_{t-j}^{\mathrm{eff}} - y_{t-j-1}^{\mathrm{eff}}\right)$.
To ensure comparability across series with very different demand levels, all target-based features (including differences and the slope proxy) are subsequently normalized by a series-specific scale factor defined from the recent 53-week effective demand; details are provided in the paragraph about Scaling~\ref{sec:scaling}.

Seasonality is captured using week-of-year as a categorical feature and Fourier terms (sine/cosine pairs for the first three harmonics), complemented by a ``last-year window'' feature that averages the same period from the previous year. Additionally, a seasonality-strength proxy is computed from correlations between current sales and seasonal lags to distinguish series with a stable periodic structure. 

Intermittency and spikes are handled via robust statistics: a rolling median and MAD yield a robust z-score that flags spike-like weeks. 
For highly intermittent items, the rolling median can be zero over the window, so the z-score should be interpreted primarily as an indicator of a non-zero demand event rather than a magnitude-based outlier measure.  This behavior is intentional: the derived \texttt{time\_since\_spike} feature then acts as a proxy for the time since the last non-zero sale, while the rolling non-zero rate complements it by distinguishing truly sparse series from low but regular demand. 

Feature importance is computed separately for each horizon-specific estimator ($h\in\{1,2,3\}$), since each model is trained on a different target shift, because of direct approach. Importances are obtained with CatBoost \texttt{LossFunctionChange} evaluated on the training \texttt{Pool} (with observation weights and categorical variables).
For readability, raw LFC scores are converted into within-horizon percentages. Table~\ref{tab:fi_by_horizon} lists selected high-impact drivers; therefore, the displayed percentages do not necessarily sum to 100 within a column.
To keep feature names interpretable, Fourier seasonality terms are grouped by harmonic order ($k=1,2,3$).

\begin{table}[t]
\centering
\small
\setlength{\tabcolsep}{5pt}
\begin{tabular}{l r r r}
\toprule
Feature & $h=1$ (\%) & $h=2$ (\%) & $h=3$ (\%) \\
\midrule
\texttt{week\_of\_year}                    & 22.48 & 36.15 & 35.19 \\
\texttt{fourier\_sin\_2}                   &  4.43 &  8.84 &  9.02 \\
\texttt{Store}                             &  3.53 &  8.45 &  4.95 \\
\texttt{fourier\_sin\_3}                   &  4.69 &  4.06 &  4.73 \\
\texttt{Product}                           &  1.99 &  3.72 &  3.84 \\
\texttt{unique\_id}                        &  1.17 &  3.33 &  1.71 \\
\texttt{seasonality\_strength}             &  3.17 &  2.15 &  3.18 \\
\texttt{fourier\_cos\_3}                   &  1.61 &  1.36 &  3.20 \\
\texttt{fourier\_sin\_1}                   &  1.50 &  1.10 &  2.76 \\
\texttt{fourier\_cos\_1}                   &  2.05 &  0.70 &  2.47 \\
\texttt{fourier\_cos\_2}                   &  2.39 &  2.29 &  2.40 \\
\texttt{lag\_51}                           &  3.18 &  0.48 &  0.30 \\
\texttt{std\_8}                            &  2.69 &  1.71 &  0.69 \\
\texttt{iqr\_13}                           &  2.62 &  0.56 &  0.34 \\
\texttt{lag\_0}                            &  2.47 &  0.55 &  0.33 \\
\texttt{last\_year\_window}                &  2.46 &  2.78 &  0.41 \\
\texttt{time\_since\_spike}                &  2.40 &  1.71 &  1.71 \\
\texttt{momentum\_5}                       &  1.86 &  2.43 &  0.19 \\
\texttt{nonzero\_rate\_12}                 &  1.53 &  0.98 &  1.59 \\
\bottomrule
\end{tabular}
\caption{Selected features with the highest CatBoost \texttt{LossFunctionChange} importance. Values are expressed as within-horizon percentages derived from raw LFC scores; only a subset of features is shown, so column totals are below 100.}
\label{tab:fi_by_horizon}
\end{table}

\paragraph{Scaling.}
\label{sec:scaling}
A key difficulty in global modeling is scale heterogeneity: some store--product pairs sell fractions of a unit per week on average, while others sell orders of magnitude more. Training a global model on raw time series is primarily influenced by high-scale ones, because their errors dominate the loss. In a global setting, this shifts the model’s capacity toward explaining differences in level across items rather than capturing within-series temporal dynamics. As a result, the model tends to prioritize fitting the overall demand level rather than learning patterns such as trends or seasonality. Applying scaling mitigates this imbalance and encourages the estimator to focus on the series' shape (patterns) rather than its mean level. A per-series dynamic scale factor\footnote{The initial idea for this annualized scaling heuristic was suggested by Adrian Foltyn, Data Science Manager at Nestl\'e (personal communication, May 2022), and was later adapted and extended by the author for the VN2 competition.} is computed as the 53-week rolling mean over in-stock weeks (backfilled for early weeks), multiplied by 53 and clipped to a minimum of 1.

\[
\textit{scale\_factor}^{(i)}_{t}
=
\max\!\Bigl( 53 \cdot \operatorname{mean}^{\mathrm{non\text{-}missing}}\!\bigl( y^{(i)}_{\mathrm{eff},\,t-52:t} \bigr),\ 1 \Bigr).
\]

\noindent where $\operatorname{mean}^{\mathrm{non\text{-}missing}}(\cdot)$ denotes the average computed over available (non-missing) values within the specified window.

Intuitively, this approximates an annualized demand level and provides a robust normalization constant. To avoid undefined normalization for late series starts and new product introductions, the 53-week estimate is used only when sufficient effective history is available (at least 45 in-stock observations). Earlier points rely on a warm-start proxy based on the expanding mean of $y^{\mathrm{eff}}$ (annualized to 53 weeks and clipped to at least 1), ensuring that the scale factor and the resulting forecasts remain well-defined even when the observed history is shorter than one year. In production settings with extremely sparse early histories or new product introductions, this warm-start can be complemented by a cross-sectional prior at higher hierarchy levels (e.g., product- or store-level typical scale) before enough series-specific observations accumulate. During fitting, the horizon-specific targets and target-based numeric features (lags and rolling statistics) are divided by this scale factor. At prediction time, forecasts are rescaled back to the original units. For example, a 5-unit uplift has very different implications for a series with a baseline of 2 units/week than for one with a baseline of 80 units/week; scaling maps both series into a comparable learning space where the model can focus on \emph{relative} changes and patterns instead of absolute level. This improves the stability of the global estimator and reduces the need for per-series calibration. Because masking and rolling windows can introduce missing feature values, I applied a two-level median imputation strategy: missing values are first filled using per-series medians of scaled values (computed on the training set), and any remaining missing values are filled with global scaled medians. While CatBoost can handle missing numerical values, the NaNs in this modeling pipeline are structural, introduced by stockout masking and rolling/window-based feature construction, rather than representing random absence. Relying on CatBoost’s native missing-value handling would encourage the estimator to exploit missingness-driven splits (treating NaNs as a special extreme) instead of learning stable demand signals. Therefore, engineered history-based features are imputed to keep their scale consistent, and well-controlled across series. This imputation does not introduce data leakage: it is applied only to the earliest parts of each time series, and these oldest observations are excluded from validation and testing by the chronological split.

\paragraph{Observation weighting.}
Retail demand is often non-stationary: trends shift, seasonality can drift, and intermittency patterns may change. Such changes are a natural consequence of how retail businesses operate. The approach does not assume stationarity of raw demand levels. Rather than applying explicit target differencing, the modeling pipeline relies on dynamic scaling and on change-based variables. Concretely, the target and all target-based features are expressed on a comparable scale using the per-series scale factor defined in the Scaling subsection ~\ref{sec:scaling}. This normalization anchors each series to a stable annualized reference estimated from recent in-stock history, reducing scale-driven dominance during fitting even when the series exhibits trends or level shifts. Differencing is therefore used as an input signal (trend/momentum variables), while the model remains trained to predict demand in the scaled units. However, portfolio evolution, substitution, and cannibalization within product groups, and rare but impactful events like COVID-19 can all shift the underlying demand-generating process. Rather than relying on frequent retraining or heavy-handed truncation of history, time-decayed observation weights are applied to emphasize recent data and learn the most up-to-date behaviors. Specifically, each training row receives a weight based on its recency within the series, with stepwise decay across yearly blocks (highest weight for the most recent year and progressively lower weights for older years). In practice, weights are assigned per series in yearly ($\approx$ 53-week) blocks: the most recent block has a weight of 1.0, and each preceding block is multiplied by 0.5 relative to the block that follows (example: 1.0 → 0.5 → 0.25). The decay factor can be treated as a hyperparameter controlling how strongly the model prioritizes recent observations. This simple weighting scheme pushes the model toward the most relevant patterns, while still leveraging a longer history for seasonal and behavioral characterization. Observation weighting, like the iterative feature additions, was introduced based on validation error analysis and retained because it improved performance compared with uniform weights.

In practice, the combination of (i) a global model that pools information across series, (ii) dynamic scaling that mitigates cross-series magnitude imbalance, and (iii) recency weighting that adapts to behavioral drift, provides a strong and robust foundation for multi-horizon retail forecasting in the VN2 setting.

\section{Ordering Policy}
\label{sec:ordering_policy}
This section describes the ordering policy that maps the information available at the end of week $t$ into an order decision $Q_t^{(i)}$ for each item $i$. The policy is designed to be consistent with the problem setup in Section~\ref{sec:problem}: periodic review, two-week lead time (orders placed at the end of $t$ are available at the start of $t{+}3$), and the objective that penalizes both shortages ($c_s L_t^{(i)}$) and end-of-week inventory ($c_h E_t^{(i)}$). The resulting policy is intentionally simple and transparent: it combines multi-horizon point forecasts with an explicit inventory projection to the delivery week and a cost-aware target stock computation.

\subsection{Forecast Generation}
\label{sec:forecast_generation}
At decision time (end of week $t$), the forecasting model (Section~\ref{sec:forecasting_model}) produces point forecasts for each item $i$ for the next three weeks:
\[
\hat{D}_{t+1}^{(i)}, \ \hat{D}_{t+2}^{(i)}, \ \hat{D}_{t+3}^{(i)}.
\]
While the ordering decision $Q_t^{(i)}$ affects availability in week $t{+}3$, the intermediate forecasts for $t{+}1$ and $t{+}2$ are required to project the inventory position through the lead time. Forecasts are post-processed to respect the discrete and non-negative nature of weekly unit demand:
\[
\hat{D}_{t+h}^{(i)} \leftarrow \max\left(\mathrm{round}(\hat{D}_{t+h}^{(i)}),\, 0\right), \quad h \in \{1,2,3\}.
\]

\subsection{Project inventory to the start of week $t{+}3$}
\label{sec:inventory_projection}
At the end of week $t$, the observed system state provides the end-of-week inventory $E_t^{(i)}$ and the transit quantities that will be received at the start of $t{+}1$ and $t{+}2$, denoted by $R_{t+1}^{(i)}$ and $R_{t+2}^{(i)}$ (these correspond to earlier orders, per the definition of $R_t^{(i)}$ in Section~\ref{sec:problem}). Since the setting assumes lost sales and inventory cannot become negative, the evolution of inventory during weeks $t{+}1$ and $t{+}2$ is approximated by simulating sales using the corresponding point forecasts.

First, the start-of-week inventory at $t{+}1$ is computed:
\[
I_{t+1}^{(i)} = E_t^{(i)} + R_{t+1}^{(i)}.
\]
Then, sales in week $t{+}1$ is approximated using $\hat{D}_{t+1}^{(i)}$, which yields the projected end-of-week inventory:
\[
\tilde{E}_{t+1}^{(i)} = \max\left(I_{t+1}^{(i)} - \hat{D}_{t+1}^{(i)},\, 0\right).
\]
Analogously for week $t{+}2$:
\[
I_{t+2}^{(i)} = \tilde{E}_{t+1}^{(i)} + R_{t+2}^{(i)}, \qquad
\tilde{E}_{t+2}^{(i)} = \max\left(I_{t+2}^{(i)} - \hat{D}_{t+2}^{(i)},\, 0\right).
\]
Finally, the projected inventory available at the \emph{start} of week $t{+}3$ (before any sales in $t{+}3$ occur) is
\[
\tilde{I}_{t+3}^{(i)} = \tilde{E}_{t+2}^{(i)}.
\]
This quantity represents the inventory position that would be available in the first week affected by the order $Q_t^{(i)}$, under a point-forecast simulation of demand and the lost-sales assumption \citep{lariviere1999stalking}.

\subsection{Cost-aware target stock for week $t{+}3$}
\label{sec:target_stock}
Given the projected start-of-week inventory $\tilde{I}_{t+3}^{(i)}$, the policy computes a target stock level for week $t{+}3$ that reflects the trade-off between shortage and holding costs. Under the classical single-period newsvendor approximation (linear shortage and holding costs, and a specified demand distribution), the cost-minimizing service level is: \citep{zipkin2000foundations,silver1998inventory}
\[
q^\star = \frac{c_s}{c_s + c_h},
\]
which in VN2 yields $q^\star = \frac{1.0}{1.0+0.2} \approx 0.8333$. This $q^\star$ is a simple and widely used rule from the newsvendor setting: it converts the cost trade-off into a target service level. When the shortage cost $c_s$ is high relative to the holding cost $c_h$, the policy aims for a higher service level (more safety stock). A useful sanity check is $c_s=c_h$, which gives $q^\star=0.5$: under- and over-stocking are equally costly, so there is no reason to systematically bias the target stock upward or downward.
This service level is mapped to a standard normal quantile:
\[
z_q = \Phi^{-1}(q^\star) \approx 0.9674,
\]
where $\Phi^{-1}$ is the inverse CDF of the standard normal distribution. This maps the service level to a safety factor under a normal-approximation heuristic; in VN2, it is used as a lightweight rule rather than a claim of exact optimality under the full multi-period dynamics. Note that the normal mapping is a lightweight convenience: retail errors, especially for intermittent items, are rarely Gaussian. In production, $q^\star$ would be implemented via empirical quantiles or a probabilistic forecast, avoiding any normality assumption. In the symmetric case $q^\star=0.5$, the corresponding normal quantile is $z_q=0$, so the safety correction disappears and the target reduces to the point forecast.

Here $z_q$ plays the role of a cost-driven safety factor expressed in units of standard deviation: it says how many ``sigmas'' above the point forecast the target stock should be. Since the model produces point forecasts rather than a full predictive distribution, the uncertainty scale is approximated with a lightweight proxy based on the level of the week-$t{+}3$ forecast:
\[
\sigma_{t+3}^{(i)} = \phi \sqrt{\hat{D}_{t+3}^{(i)}}
\]
The square-root dependence means the buffer increases with demand level. The scalar $\phi$ is calibrated on the same time-based validation window as the estimator’s hyperparameter optimization. The objective is to minimize the total cost, defined as the sum of shortage and holding costs, using forecasts generated for that validation period, so the calibration naturally reflects forecast errors. In production settings, the validation window should reflect the operating conditions (e.g., including relevant seasonal periods), as the error distribution and inventory risk can vary throughout the year.

The resulting cost-aware target stock for week $t{+}3$ is then:
\[
B_{t+3}^{(i)} = \hat{D}_{t+3}^{(i)} + z_q \, \sigma_{t+3}^{(i)}.
\]
Intuitively, $B_{t+3}^{(i)}$ corresponds to the point forecast augmented by a safety buffer that increases when shortage cost dominates holding cost (via $z_q$) and when forecast magnitude suggests higher variability (via $\sigma_{t+3}^{(i)}$). With a single global $\phi$, $z_q$ and $\phi$ could be merged into one calibration term. They are kept separate on purpose: $z_q$ is cost-driven (it comes from $c_s$ and $c_h$), while $\phi$ captures how much extra buffer is needed based on data patterns. This also keeps the door open to extend $\phi$ later (e.g., by horizon, store, product, or segment) without changing the overall architecture structure. In practice, this flexibility helps avoid policies that look good on the global cost metric but behave poorly for a small set of SKUs, since $\phi$ can be tuned more conservatively or segmented when needed.

\subsection{Order rule}
\label{sec:order_rule}
The order placed at the end of week $t$ is chosen to raise the projected inventory at the start of week $t{+}3$ up to the target stock level:
\[
Q_t^{(i)}=\max\!\left(\left[\,B_{t+3}^{(i)}-\tilde I_{t+3}^{(i)}\,\right],\,0\right).
\]
The max operator enforces non-negativity, and in practice, for many store–product combinations, this value is exactly zero, as desired. If projected stock is already above the target, the best decision is not to order anything.

\section{Conclusion}
\label{sec:conclusion}
This report presented the winning VN2 solution as an end-to-end \textit{predict-then-optimize} pipeline built around a single global multi-horizon forecasting model and a lightweight, cost-aware ordering policy. The key takeaway is that strong performance in inventory planning does not necessarily require a complex decision model when the objective is explicitly defined and the information structure is stable. Here, the objective is provided as linear holding and shortage penalties; in practice, these costs are rarely observed directly and depend on many business factors, so the effective holding and shortage costs can vary by product and even by store–product combination. Instead, the combination of (i) a carefully engineered global forecaster that can handle heterogeneous behaviors, intermittency, and stockout-censored histories, and (ii) a transparent policy that projects inventory to the delivery week and converts forecasts into cost-aligned target stock levels, is sufficient to achieve a robust and scalable solution.

From a forecasting perspective, the results reinforce a practical lesson: the estimator is only one component of the system, and feature engineering is the bridge between real-world demand behavior and a global model's ability to generalize. Stockout-aware feature construction, dynamic per-series scaling to reduce cross-series magnitude imbalance, and time-decayed sample weighting to adapt to drift together enable a single CatBoost-based approach to cover a wide range of demand regimes. From a decision-making perspective, the policy demonstrates that explicitly incorporating lead time and asymmetric costs can be more impactful than increasing model complexity. The derived service-level intuition provides a simple yet effective mechanism for trading off shortage and holding penalties in a manner consistent with the evaluation metric.

Several extensions could further improve the methodology and broaden applicability beyond the VN2 setup. First, the ordering stage could be upgraded from point-forecast buffering to \textit{probabilistic} decision-making by learning (or approximating) predictive distributions and directly computing quantiles, rather than relying on a hand-crafted uncertainty proxy. Second, the cost-aware rule could be generalized to incorporate additional operational constraints that are common in retail, such as case-pack rounding, minimum order quantities, capacity limits, and cross-item coupling (e.g., shared shelf space or budget constraints). Third, demand censoring could be addressed more explicitly via models that infer latent demand under stockouts (e.g., censored regression or imputation schemes), potentially improving both forecasting accuracy and downstream ordering decisions for fast movers. Fourth, the validation protocol could be expanded to stress-test robustness under regime shifts and to jointly tune policy parameters and forecasting hyperparameters using a simulator-based objective. Finally, while this report intentionally focused on a modular two-stage design, an interesting research direction is to explore hybrid approaches that preserve interpretability while optimizing end-to-end cost, for example by learning a small residual correction on top of the analytical policy or by using reinforcement learning constrained by a base-stock prior.

The practical value of the proposed approach can be quantified against the VN2 benchmark, which combines a seasonally weighted 13-week moving-average forecast with a 4-week coverage policy. The resulting total cost is reduced by 13.2\% (3.763€ vs. 4.334€). Overall, the VN2 benchmark provides a compact yet realistic environment where forecasting quality and operational decision-making are tightly coupled. The proposed solution demonstrates that a principled, modular design—global learning for demand patterns and an explicit cost-aware mapping for replenishment—can deliver strong performance while remaining simple to implement, analyze, and extend.

\bibliographystyle{plainnat}
\bibliography{references}

\end{document}